# A Backwards View for Assessment

Ross D. Shachter and David E. Heckerman
Stanford University
Stanford, California  94305-4025

## Introduction

Much artificial intelligence research focuses on the problem of deducing the validity of unobservable propositions or *hypotheses* from observable *evidence*.[1] Many of the knowledge representation techniques designed for this problem encode the relationship between evidence and hypothesis in a directed manner. Moreover, the direction in which evidence is stored is typically *from evidence to hypothesis*. For example, in the rule-based approach, knowledge is represented as rules of the form:

IF <evidence> THEN <hypothesis>

In this scheme, observable propositions are most often found in the antecedent of rules while unobservable propositions are usually found in the consequent.

In early applications of these methodologies, the relationship between evidence and hypothesis was assumed to be categorical. However, as artificial intelligence researchers began to address real-world problems, they recognized that the relationship between evidence and hypothesis is often uncertain and developed methods for reasoning under uncertainty. For example, the creators of one of the earliest expert systems, MYCIN, augmented the rule-based approach to knowledge representation and manipulation with a method for managing uncertainty known as the certainty factor (CF) model [Shortliffe 75]. In this and other approaches [1, 2], the direction of knowledge representation was carried over from their categorical counterparts. In the MYCIN certainty factor model, for example, knowledge is stored in rules of the form:

IF <evidence> THEN <hypothesis>, CF

where CF is a measure of the change in belief in the hypothesis given the evidence.

That knowledge is encoded in the direction from evidence to hypothesis in such popular approaches is not accidental. This direction corresponds to the direction in which knowledge is *used* by the program to deduce or infer the validity of unobservable propositions. This appears to be convenient for problem solving. However, in this paper, we argue that in most real-world applications there are advantages to representing knowledge in the direction *opposite* to the usage direction. In particular, we argue that representing knowledge in the direction from unobservable hypothesis to observable evidence is often cognitively simpler and more efficient.

The argument is based on three observations. The first is that many real-world problems involve *causal* interactions. In this paper, we make no attempt to define causality in terms of more basic concepts; we take it to be a primitive notion. The second observation is that the direction of causality is most often opposite to that of the direction of usage. That is, *hypotheses tend to cause evidence*. There are many examples of this in medicine in which the unobservable hypothesis is the true cause of an illness and the observable evidence is the illness' effect in the form of symptoms. Of course, there are exceptions. In trauma cases such as such as automobile accidents, for example, the cause is observable and some of the effects are difficult to observe. However, tests then performed to determine the hidden effects of the accident fit the more usual model of unobservable cause and observable effect. The final observation is that people are more comfortable when their beliefs are elicited in the causal direction. It appears that it is cognitively simpler to construct assessments in that direction [3]. Furthermore, models are often less complex in the causal direction. Thus, these three observations suggest that there may be advantages to representing knowledge in the direction opposite to the direction of usage. This is summarized in Figure 1.

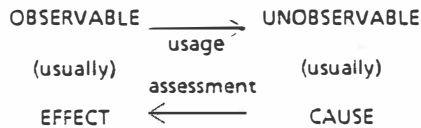

Figure 1:  The two directions of representation

## Influence Diagrams

In this paper, we will examine these issues in the context of probability theory. Nonetheless, we believe that the distinction between direction of usage and direction of natural assessment is a fundamental issue, independent of the language in which belief is represented.

Within the theory of probability, there are several graphical representations for uncertainty featuring both directed graphs [4, 5, 6, 7] and undirected graphs [8, 9]. The different representations all share the basic concept of the factorization of an underlying joint distribution, and the explicit revelation of conditional independence. We will use a directed graph representation method since directionality is central to our discussion. In particular, will use the influence diagram representation scheme [5].

Each of the oval nodes in an influence diagram represents a random variable or uncertain quantity. The arcs indicate *conditional dependence* with the successor's probability distribution conditioned on the outcomes of its predecessors. For example, there are three possible influence diagrams for the two random variables X and Y shown in Figure 2. In the first case, X has no predecessors so we assess a marginal (unconditional) distribution for X and a conditional distribution for Y given X. In the next case, with the arc reversed, we assess a marginal distribution for Y and a conditional distribution for X given Y. Both correspond to the same fundamental model, at the underlying joint distribution, but the two diagrams represent two different ways of factoring the model. The transformation between them, which involves reversing the direction of conditioning and hence the reversal of the arc, is simply Bayes' Theorem. Finally, in the third case, neither node has any predecessors so X and Y are independent. Therefore, we can obtain the joint by assessing marginal distributions for both X and Y. When the two variables are independent, we are free to use any of the three forms, but we prefer the last one which explicitly reveals that independence in the graph.

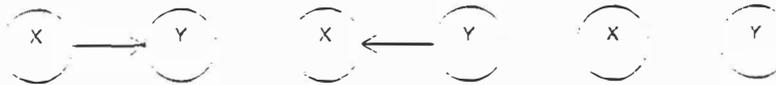

**Figure 2:** An influence diagram with two nodes

In Figure 3, we see four possible influence diagrams for the random variables, X, Y and Z. In the first case, the general situation, the three variables appear to be completely dependent. We assess a marginal distribution for X, and conditional distributions for Y given X and for Z given X and Y. In general there are $n!$ factorizations of a joint distribution among $n$ variables. Each possible permutation leads to a different influence diagram. In the second case in the figure, the three variables are completely independent; in the third case, X and Y are dependent but Z is independent of both of them. In the fourth case, we see *conditional independence*. The absence of an arc from X to Z, indicates that while X and Z are dependent, they are independent given Y. This type of conditional independence is an important simplifying assumption for the construction and assessment of models of uncertainty.

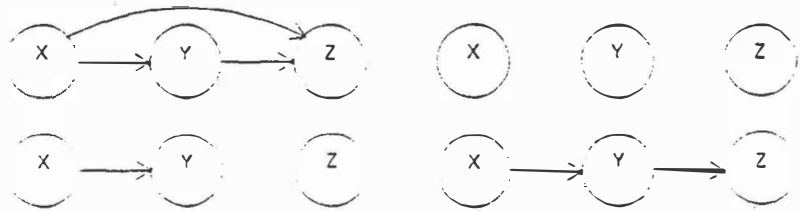

**Figure 3:** An influence diagram with three nodes

In the influence diagram we always require that there be no directed cycles. By doing so, there is always an ordering among the variables so that we can recover the underlying joint distribution. While there are some anomalous cases with cycles where the joint is still recoverable, we will require that the diagram remain acyclic.

There is one other type of influence diagram node relevant to our discussion, a deterministic node, drawn as a double oval (as opposed to the probabilistic node which we have shown as a single oval). The deterministic variable is a function of its predecessors, so its outcome is known with certainty if we observe the outcomes of those predecessors. In general, we can not observe all of those predecessors, so there can be uncertainty in the outcome of a deterministic variable.

An operation central to our discussion is the transformation of an influence diagram in the assessment direction to one in the usage direction. This is accomplished by *reversing the arcs* in the influence diagram which corresponds to the general version of Bayes' Theorem [5, 10, 11]. This is shown in Figure 4. As long as there is no other path between chance nodes (either probabilistic or deterministic node) X and Y, we can reverse the arc from X to Y. If there were another path, the newly reversed arc would create a cycle. In the process of the reversal, both chance nodes inherit each other's predecessors. This may add a considerable number of arcs to the influence diagram. We shall be concerned in later sections about the relative complexity of the model before and after reversals. Sometimes, not all of the arcs are needed after reversals, but in general they may be. If multiple reversals are needed, then the order in which they are performed may affect this arc structure.[2]

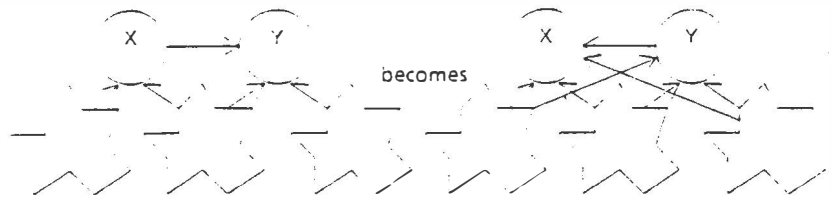

Figure 4: Arc reversal in an influence diagram

## Deterministic models

The importance of distinction between direction of assessment and direction of usage appears even in the simplest categorical models. Suppose, for example, that we have an error in the output from a small computer program. If we knew the source of the error, then we would know with certainty the type of output error to expect. Thus we could use the model shown in Figure 5 in which the programming error is represented by a probabilistic node conditioning the computer output represented by a deterministic node. When we observe the output and wish to learn about the source of the error, we reverse the arc using Bayes' Theorem and find that, after the reversal, both nodes have become probabilistic. Given the output, we do not necessarily know what type of error caused it, but we are able to update our previous beliefs about the possible errors in the light of this new evidence. Our direction of usage is clearly in the more complex, reverse direction, but the model is much more easily constructed in the original direction, which exploits the categorical behavior of our man-made computer system.

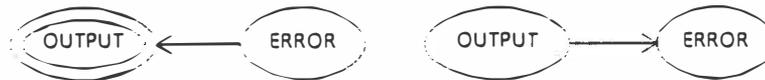

Figure 5: An influence diagram for diagnosing computer programming errors

Suppose now that we have a much larger computer system. If it were written with a modular design, we may have the influence diagram shown in Figure 6. Again this model is relatively easy to construct because of the categorical nature and the independence among subsystems. If however, we observe the output and wish to update our knowledge about those subsystems, we find that they are no longer categorical, nor are they independent, in light of the new information.[3] This newer, more complex model is the correct one to use to update our beliefs as we observe evidence, but it is less convenient for assessment.

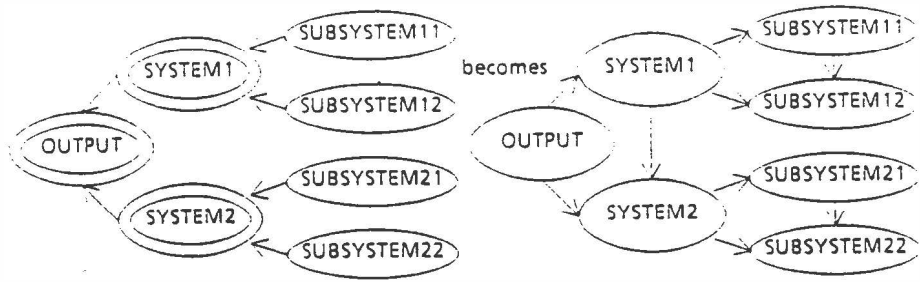

Figure 6: An influence diagram for a modular computer program

## Probabilistic models

In most real-world domains, we do not have a categorical model to assess, but there is still considerable advantage to thinking about a problem in the causal direction. Often, basic and straightforward probabilistic models become much more complex when viewed in the direction of usage.

Consider the case of two effects with a single common cause. Even when the effects are conditionally independent given the cause as in Figure 7, they are, in general, dependent when the problem is reversed. Similarly, when there are two independent causes and a single common effect as in Figure 8, we see complete dependency when the problem is reversed. Clearly as the number of causes and effects increase, the problem stays straightforward in the causal direction, but becomes even more complex in the direction of usage.

As an example, consider two disorders, congestive heart failure and nephrotic syndrome (a kidney disease), which essentially arise independently. Congestive heart failure often results in an enlarged heart (cardiomegaly) and accumulation of fluid in the ankles (pitting edema), while nephrotic syndrome often leads to protein in the urine and pitting edema as well. A simple test for protein in the urine is whether the urine is frothy, and an x-ray is used to detect cardiomegaly. The model corresponding to this is shown on the right side of Figure 9. If we turn the model around to show how the unobservable events of interest, heart failure and nephrotic syndrome, depend on the observables, x-ray, pitting edema, and frothy urine, then the model becomes the one shown on the left side of Figure 9. The original model was not only simpler, but more natural to assess, going in the causal direction. The reversed model would be intolerably confusing to assess, but it has all of the dependencies one needs for proper usage.

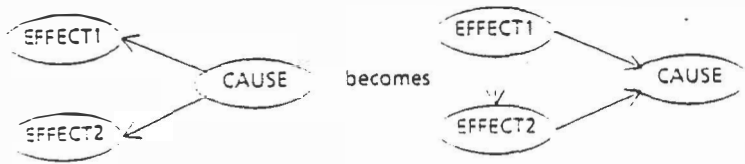

Figure 7: A single cause with multiple effects

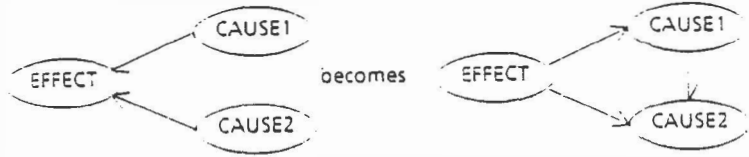

Figure 8: A single effect with multiple causes

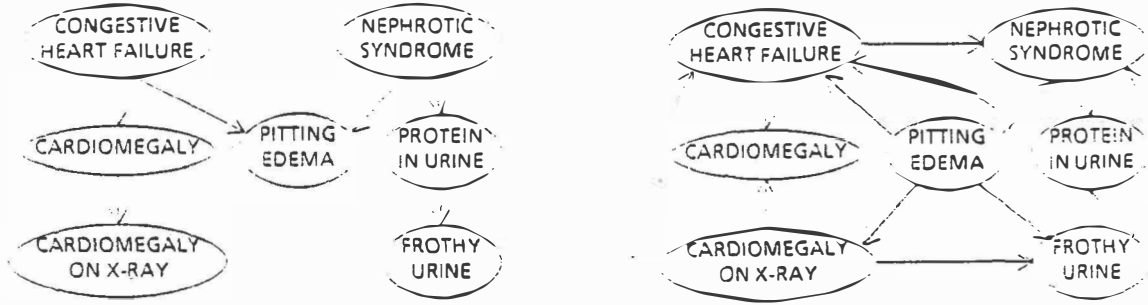

Figure 9: An influence diagram for a medical diagnosis problem

One other major advantage of viewing a problem in different directions for construction and solution is that parts of the assessment may not vary much from case to case. Consider the simple model in Figure 10, in which a probabilistic effect depends on a probabilistic cause. Often in medicine and other domains, the likelihood distribution for the effect given the cause is independent of patient-specific factors; e.g., a particular disorder will have a distribution over possible symptoms. On the other hand, the probability distribution for the cause, e.g., the distribution of possible disorders, may vary widely between cases, on the basis of age, occupation, climate and so on. When the model is reversed to the direction of usage, e.g., diagnosis for disorder given symptoms, both probabilities have become patient-specific. This suggests considerable time-saving in the construction of models by building them in the direction of natural assessment to exploit any constant likelihoods.

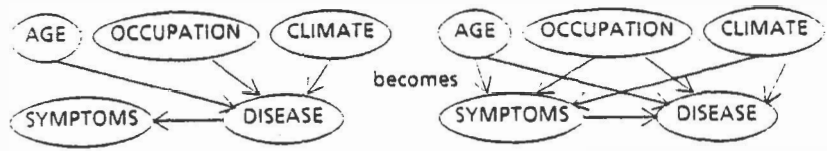

Figure 10: Taking advantage of constant likelihood

Finally, it is often useful to add new variables which simplify the construction process. Consider the medical example in Figure 9. If we are interested in the probability distribution for pitting edema given congestive heart failure, it is much easier to assess with nephrotic syndrome present in the model. We can then take expectation over nephrotic syndrome to arrive at our desired conditional distribution. If we did not explicitly consider nephrotic syndrome, then we would be forced to perform the integration mentally instead. The addition of variables can be of considerable cognitive aid when trying to assess the probability distributions. This process is what Tribus [12] calls "the law of the extension of the conversation." We get little benefit from this technique unless we first build our model in the causal direction.

## Conclusions

We believe it is important to distinguish between the direction in which a model is constructed and the direction in which it will be applied. Models which capture the richness of the interactions among their variables can become impossible to assess in the usage direction, even though they may be simple and natural to think about and assess in the opposite direction.

Artificial intelligence researchers have argued that various methods for reasoning with uncertainty are impractical because of the complexity of knowledge assessment and belief revision [13, 14]. Indeed, many AI researchers have sacrificed self-consistency of the reasoning mechanism in order to facilitate simplicity in the knowledge representation process [15]. We contend that the desired simplicity can be found often by constructing models in the direction opposite that of usage without having to sacrifice fundamentals.

## Acknowledgements

This work was supported in part by Decision Focus, Inc., the Josiah Macy, Jr. Foundation, the Henry J. Kaiser Family Foundation, and the Ford Aerospace Corporation. Computing facilities were provided by the SUMEX-AIM resource under NIH grant RR-00785.

## Notes

[1] Although different meanings have been attached to the terms "evidence" and "hypothesis," we will take "hypothesis" to mean an unobservable proposition and a item of "evidence" to mean an observable proposition.

[2] It is interesting to note that the reversal operation simplifies considerably when the predecessor X is a deterministic node. It does not simplify, however, when the successor Y is the deterministic node.

[3] Notice that the newer model does show some conditional independence which can be exploited at the time of usage.

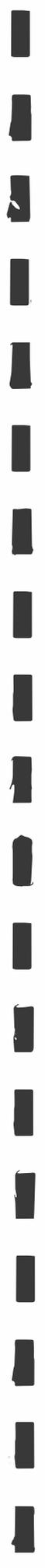